# A CHAID Based Performance Prediction Model in Educational Data Mining

**M. Ramaswami[1] and R. Bhaskaran[2]**

**[1] Department of Computer Science(DDE), Madurai Kamaraj University
Madurai, Tamilnadu 625 021, India**

**[2] School of Mathematics, Madurai Kamaraj University
Madurai, Tamilnadu 625 021, India**

## Abstract

The performance in higher secondary school education in India is a turning point in the academic lives of all students. As this academic performance is influenced by many factors, it is essential to develop predictive data mining model for students' performance so as to identify the slow learners and study the influence of the dominant factors on their academic performance. In the present investigation, a survey cum experimental methodology was adopted to generate a database and it was constructed from a primary and a secondary source. While the primary data was collected from the regular students, the secondary data was gathered from the school and office of the Chief Educational Officer (CEO). A total of 1000 datasets of the year 2006 from five different schools in three different districts of Tamilnadu were collected. The raw data was preprocessed in terms of filling up missing values, transforming values in one form into another and relevant attribute/ variable selection. As a result, we had 772 student records, which were used for CHAID prediction model construction. A set of prediction rules were extracted from CHIAD prediction model and the efficiency of the generated CHIAD prediction model was found. The accuracy of the present model was compared with other model and it has been found to be satisfactory.

## 1. Introduction

Students' academic performance hinges on diverse factors like personal, socio-economic, psychological and other environmental variables. Prediction models that include all these variables are necessitated for the effective prediction of the performance of the students. The prediction of student performance with high accuracy is beneficial to identify the students with low academic achievements initially. The identified students can be individually assisted by the educators so that their performance is better in future. In these perspectives, a number of data mining models ([4], [5], [6]) and statistical models ([2], [3]) have been constructed with inputs of varied factors that influence the performance of the students. It was reported that models constructed through data mining of inductive exercise were better in terms of prediction accuracy than those constructed through statistical measures with hypothetico-dedeuctive approach [15]. As data mining models have relatively higher degree of accuracy, we use such tools to develop predictive data mining model for students' performance in Indian educational system.

School education in India is a two-tier system, the first ten years covering general education followed by two years of senior secondary education. This two-year education, which is also known as Higher Secondary Education, is important because it is a deciding factor for opting desired subjects of study in higher education. In fact, the higher secondary education acts as a bridge between school education and the higher learning specializations that are offered by colleges and universities. In this connection, the objectives of the present investigation were framed so as to assist the low achievers in higher secondary level and they are

- (a) Generation of a data source of predictive variables,
- (b) Identification of highly influencing predictive variables on the academic performance of higher secondary students
- (c) Construction of a CHAID prediction model in data mining on the basis of identified predictive variables and
- (d) Validation of the developed CHAID prediction model for higher secondary students studying in Indian educational system.

## 2. Previous Studies

A number of reviews pertaining to not only the diverse factors like personal, socio-economic, psychological and other environmental variables that influence the performance of students but also the models that have been used for the performance prediction are available in the literature and a few specific studies are listed below for reference.

Walters and Soyibo [1] conducted a study to determine Jamaican high school students' (population n=305) level of performance on five integrated science process skills with performance linked to gender, grade level, school location, school type, student type, and socio-economic background (SEB). The results revealed that there was a positive significant relationship between academic performance of the student and the nature of the school.

Khan [2] conducted a performance study on 400 students comprising 200 boys and 200 girls selected from the senior secondary school of Aligarh Muslim University, Aligarh, India with a main objective to establish the prognostic value of different measures of cognition, personality and demographic variables for success at higher secondary level in science stream. The selection was based on cluster sampling technique in which the entire population of interest was divided into groups, or clusters, and a random sample of these clusters was selected for further analyses. It was found that girls with high socio-economic status had relatively higher academic achievement in science stream and boys with low socio-economic status had relatively higher academic achievement in general.

Hijazi and Naqvi [3] conducted as study on the student performance by selecting a sample of 300 students (225 males, 75 females) from a group of colleges affiliated to Punjab university of Pakistan. The hypothesis that was stated as "Student's attitude towards attendance in class, hours spent in study on daily basis after college, students' family income, students' mother's age and mother's education are significantly related with student performance" was framed. By means of simple linear regression analysis, it was found that the factors like mother's education and student's family income were highly correlated with the student academic performance.

Kristjansson, Sigfusdottir and Allegrante [7] made a study to estimate the relationship between health behaviors, body mass index (BMI), self-esteem and the academic achievement of adolescents. The authors analyzed survey data related to 6,346 adolescents in Iceland and it was found that the factors like lower BMI, physical activity, and good dietary habits were well associated with higher academic achievement.

Moriana et al. [8] studied the possible influence of extra-curricular activities like study-related (tutoring or private classes, computers) and/or sports-related (indoor and outdoor games) on the academic performance of the secondary school students in Spain. A total number of 222 students from 12 different schools were the samples and they were categorized into two groups as a function of student activities (both sports and academic) outside the school day. Analysis of variance (ANOVA) was used to verify the effect of extra curricular actives on the academic performance and it was observed that group involved in activities outside the school yielded better academic performance.

Bray [9], in his study on private tutoring and its implications, observed that the percentage of students receiving private tutoring in India was relatively higher than in Malaysia, Singapore, Japan, China and Srilanka. It was also observed that there was an enhancement of academic performance with the intensity of private tutoring and this variation of intensity of private tutoring depends on the collective factor namely socio-economic conditions.

Modeling of student performance at various levels is discussed in [4], [5], and [6]. Ma, Liu, Wong, Yu, and Lee [4] applied a data mining technique based on association rules to find weak tertiary school students (n= 264) of Singapore for remedial classes. Three scoring measures namely Scoring Based on Associations (SBA-score), C4.5-score and NB-score for evaluating the prediction in connection with the selection of the students for remedial classes were used with the input variables like sex, region and school performance over the past years. It was found that the predictive accuracy of SBA-score methodology was 20% higher than that of C4.5 score, NB-score methods and traditional method.

Kotsiantis, et al. [5] applied five classification algorithms namely Decision Trees, Perceptron-based Learning, Bayesian Nets, Instance-Based Learning and Rule-learning to predict the performance of computer science students from distance learning stream of Hellenic Open University, Greece. A total of 365 student records comprising several demographic variables like sex, age and marital status were used. In addition, the performance attribute namely mark in a given assignment was used as input to a binary (pass/fail) classifier. Filter based variable selection technique was used to select highly influencing variables and all the above five classification models were constructed. It was noticed that the Naïve-Bayes algorithm yielded high predictive accuracy (74%) for two-class (*pass/fail*) dataset.

Al-Radaideh, et al. [10] applied a decision tree model to predict the final grade of students who studied the C++ course in Yarmouk University, Jordan in the year 2005. They used 12 predictive variables and a 4-class response variable for the model construction. Three different classification methods namely ID3, C4.5, and the NaïveBayes were used. The outcome of their results indicated that Decision Tree model had better prediction than other models with the predictive accuracy of 38.33% for four-class response variable.

Cortez and Silva [6] attempted to predict failure in the two core classes (Mathematics and Portuguese) of two secondary school students from the Alentejo region of Portugal by utilizing 29 predictive variables. Four data mining algorithms such as Decision Tree (DT), Random Forest (RF), Neural Network (NN) and Support Vector Machine (SVM) were applied on a data set of 788 students, who appeared in 2006 examination. It was reported that DT and NN algorithms had the predictive accuracy of 93% and 91% for two-class dataset (*pass/fail*) respectively. It was also reported that both DT and NN algorithms had the predictive accuracy of 72% for a four-class dataset.

From these specific studies, we observe that the student performance could depend on diversified factors such as demographic, academic, psychological, socio-economic and other environmental factors. It was learnt that the predictive accuracy of constructed models ranged between 38.33 to 93%. The variation in the predictive accuracy could be correlated with the nature of student data set and utilization of number of records, predictive variables and class values of response variable. In addition, the predictive variation could be associated with the number of class values of response variable. Based on these observations, we constructed a CHAID prediction model with 7-class response variable by using highly influencing predictive variables obtained through feature selection technique so as to evaluate the academic achievement of students at higher secondary level in India.

## 3. Methodology

Survey cum experimental methodology was adopted to generate a database for the present study. The basis of this database was from the primary and secondary sources. While the primary data was collected from the regular students, the secondary data was gathered from the school and CEO office.

### 3.1. Data Source

A detailed questionnaire was prepared with the reference, assistance and guidance from the (i) Review of related literature, (ii) Teachers of diversified schools, (iii) Parents of higher secondary students, (iv) Joint Director of Higher Secondary Education, Government of Tamilnadu and (v) Educational experts in colleges and Universities. The designed four page A4 size close-ended questionnaire was used for the collection of student details. Most of the information of the variables was collected directly from the students through this questionnaire. Based on this information, a few derived variables were generated. While some of the information for the variables was extracted from the school records, the mark details were collected from the CEO office. All the predictor and response variables which were derived from the questionnaire are given in Table 1 for reference.

Table 1: Student Related Variables

| Variable Name | Description | Domain |
|---|---|---|
| (1)  SEX | student's sex | {*male, female*} |
| (2)  BMI | student's body mass index | {*underweight, normal weight, over weight, obesity*} |
| (3)  VAcuity | student's eye visual acuity | {*normal, defect*} |
| (4)  Comm | student's community | {*OC, BC, MBC, SC, ST*} |
| (5)  PHD | physically handicapped or not | {*yes, no*} |
| (6)  FHBT | student's food habit | {*veg , non-veg*} |
| (7)  FAM-Size | student's family size | {*one, two, three, four, five, six, seven, eight, nine, ten*} |
| (8)  LArea | student's living area | {*corporation, municipal, rural*} |
| (9)  No-EB | number of elder brothers | {*zero, one, two, three, four, five*} |
| (10) No-ES | number of elder sisters | {*zero, one, two, three, four, five*} |
| (11) No-YB | number of younger brothers | {*zero, one, two, three, four, five*} |
| (12) No-YS | number of younger  sisters | {*zero, one, two, three, four, five*} |
| (13) JIFamily | student's family status | {*individual, joint*} |
| (14) TransSchool | mode of  transportation to school | {*bicycle, city-bus, school-bus, own-vehicle, hired-vehicle,  others*} |
| (15) Veh-Home | own vehicle at home | {*no-vehicle, bi-cycle, moped, bike, car,  others*} |

| (16) PSEdu | student had primary education | {yes, no} |
|---|---|---|
| (17) ESEdu | type of institution at elementary level | {private, municipal, government} |
| (18) StMe | type of secondary syllabus | {state board, matric} |
| (19) XMark-Grade | marks/grade obtained at secondary level | {O – 90% -100%, A – 80% - 89%, B – 70% - 79%, C – 60% - 69%, D – 50% - 59%, E – 40% - 49%, F - < 40%} |
| (20) MED | medium of Instruction | {Tamil, English} |
| (21) PTution | private tuition- number of subjects | (zero, one, two, three, four, five} |
| (22) Group | group of study | {first, second, third, vocational} |
| (23) TYP-SCH | type of school | {co-education, boys, girls} |
| (24) LOC-SCH | location of school | {rural, municipal, corporation} |
| (25) SPerson | sports/athletic | {yes, no} |
| (26) SpIndoor | type of indoor game | {chess, carom, table tennis, others} |
| (27) SpOutdoor | type of outdoor game | {football, ko-ko, basketball, volleyball, kabadi, badminton, athletic, others} |
| (28) CStudy | care of study at home | {parents, grand-parents, father only, mother only, self, others} |
| (29) FEDU | father's education | {no-education, elementary, secondary, graduate, post-graduate, professional, $^\mathcal{E}$not-applicable} |
| (30) FOCC | father's occupation | {cooley, private, govt. service, business, farmer, professional, educational-institution, retired, $^\mathcal{E}$not-applicable} |
| (31) FSAL | father's monthly income | {1.. 2.4k, 2.5k .. 4.9k, 5k .. 9.9k,10k .. 14k, 15k ..25k, above 25k, $^\mathcal{E}$not-applicable} |
| (32) MEDU | mother's education | {no-education, elementary, secondary, graduate, post-graduate, professional, $^\mathcal{E}$not-applicable} |
| (33) MOCC | mother's occupation | {housewife, cooley, private, govt. service, business, professional, educational-institution, retired, $^\mathcal{E}$not-applicable} |
| (34) MSAL | mother's monthly income | {0 .. 0.9k, 1k .. 2.9k, 3k ...4.9k, 5k ..9k, 10k.. 20k, above 20k, $^\mathcal{E}$not-applicable} |
| **(Response Variable)** (35) HScGrade | marks/grade obtained at HSc Level | {O – 90% - 100%, A – 80% - 89%, B – 70% - 79%, C – 60% - 69%, D – 50% - 59%, E – 40% - 49%, F - < 40%} |

The domain values for some of the nativity variables were defined for the present investigation as follows:

- **FAM-Size-**Family Size. According to population statistics of India, the average number of children in a family is 3.1. Therefore, the maximum family size is fixed as 10 and possible range of values is from *one* to *ten*.
- **StMe-**State board/ Matric board. Two types of secondary education is offered in India through state board and matric systems. Most of the government funded schools offer state board syllabus, whereas unaided, private schools follow matric pattern. The number of subjects and examination pattern vary in both streams. Students who has matric pattern syllabus up to their secondary may perform well in the higher secondary examination. Therefore this factor may also influence the student performance at higher secondary level. Possible values are *state, matric*.
- **XMARK-Grade-**Marks obtained at secondary level. Students who is in state board stream appear for five subjects each carry 100 marks, whereas in matric stream students appear for 10 subjects each carry 100 marks. We transform the marks in percentage into grades by mapping *O – 90% to 100%, A – 80% -*

*89%, B – 70% - 79%, C – 60% - 69%, D – 50% - 59%, E – 40% - 49%,* and *F - < 40%}*.

- **BMI-**Body mass index (BMI) is a measure of body fat based on height and weight. Four possible values are fixed based on BMI value – *underweight*, *normal weight*, *overweight* and *obesity*.
- **MED-**The present study covers the educational institutions in Tamilnadu state only. Here, medium of instructions are Tamil or English.
- **Comm-**Community- Even though India has defined itself as a secular state, religion and caste are deeply entrenched in the identity of Indians across ages. These factors play a direct or indirect role in the daily lives including the education of young people. In terms of social status, the Indian population is grouped into five categories: Scheduled Castes (SC), Scheduled Tribes (ST), Most Backward Classes (MBC), Backward Classes (BC) and Others (OC). Possible values are *OC, BC, MBC, SC* and *ST*.
- **Vacuity-**Visual acuity is the eye's ability to detect fine details and it is the quantitative measure of the eye's ability to see an in-focus image at a certain distance. Possible values are *normal, defect*.
- **PTution-**Most of the parents send their wards for private tutoring after school hours. The number of subjects taught at higher secondary level is five –

part-1(Tamil), part-2(English) and part-3 (three core subjects). Most of the students prefer private tutoring for core subjects. Therefore the number private tutoring subjects can vary from *zero to five*.

- **Group-**Four types of group of study, based on core subjects, is offered at higher secondary level comprising *first* group (Mathematics, Chemistry, Physics, Biology), *second* group (Mathematics, Chemistry, Physics, Computer Science) *third* group (Mathematics, Chemistry, Physics, Computer Science), and *vocational* group (Business Mathematics, Economics, Accountancy).

- **SPerson-**This is related to the sport activities of the students, if any. The domain value of Indoor games are fixed as *chess, carom, table tennis*, and others*;* for outdoor games, the values are *football, ko-ko* (traditional game*), basketball, volleyball, kabadi* (traditional game*), badminton, athletic,* and *others*.

- **HScGrade-**Marks/Grade obtained at higher secondary level and it is declared as response variable. It is also split into seven class values: *O – 90% to 100%, A – 80% - 89%, B – 70% - 79%, C – 60% - 69%, D – 50% - 59%, E – 40% - 49%, F - < 40%*.

For the pilot study, we selected two schools in the Madurai district of Tamilnadu. While one was an urban-based, unaided and co-educational school with matric pattern of education, the other one was a rural-based, aided and co-educational school with state board pattern of education. A total of 224 (120 males, 124 females) students of higher secondary education from these two schools who appeared in April 2005 examination were the samples for our study. All the information related to student's demographic, academic and socio-economic variables was obtained from the 224 students directly through questionnaire. The mark particulars of these students were collected from the office of Chief Educational Officer after the publication of higher secondary examination results in June 2005.

A simple linear regression technique was used to construct a regression model, after encoding the categorical values of all the predictor variables into numeric values. The predictive accuracy of the student performance through this model was found to be 39.23%.

The outcome of this pilot study revealed that there was a strong correlation between the factors like location of the school, school-type, parental education and marks obtained at secondary level and the academic performance of the higher secondary students. As parental education could subdue most of the predictive variables like socio-economic condition, living area, quality of education of their children, food habits and private tutoring, the literacy rate was used as an instrument for selecting the sample schools for detailed investigation.

Based on the outcome of the pilot study, a detailed study was conducted in the subsequent higher secondary examination held in April 2006. The data sets were collected from different schools in three different districts namely Dindigul, Madurai and Kanyakumari in Tamilnadu. As per the census of the year 2001, the average literacy rate was 73.47% in Tamilnadu and the sample districts were selected on the basis of the literacy rate. While the literacy rate of Madurai was noted to be more or less the level of average literacy rate of the state, the literacy rate of Dindigul district was 68.33% (which was below the average literacy rate of the state) and the literacy rate of Kanyakumari district was 88.11% (which was above the average literacy rate of the Tamil Nadu). Five schools from these three districts each representing the factors like urban, rural, single-sex, aided and private were selected for the present investigation. A total of 1000 data sets were collected from these five different schools in these three districts. It is well known that, the representation and quality of the data is more important before constructing any predictive data mining model, we pre-processed our data set in terms of transformation, or conditioning, so that data designed to make modeling easier and more robust. As a result, we had 772 student records, which were used for CHAID prediction model construction. A total number of 228 out of 1000 student records were discarded due to the irrelevant answers by the students to the questions in the questionnaire, absence of students in the final examinations and incomplete information obtained from the school and CEO office.

## 3.2. Tools and Techniques

Classification trees are widely used in different fields such as medicine, computer science, botany and psychology [11]. These trees readily lend themselves to being displayed graphically, helping to make them easier to interpret than they would be if only a strict numerical interpretation were possible.

CHAID [12], which is one of the classification tree algorithms, is the name given to one version of the Automatic Interaction Detector that has been developed for categorical variables. In fact, CHAID is a technique that recursively partitions (or splits) a population into separate and distinct segments. These segments, called nodes, are split in such a way that the variation of the response variable (categorical) is minimized within the segments and maximized among the segments. After the initial splitting of the population into two or more nodes (defined by values of an independent or predictor variable), the splitting process is repeated on each of the nodes. Each node is treated like a new sub-population. It is then split into two or more nodes (defined by the values of another predictor variable) such that the variation of the response variable is minimized within the nodes and maximized among the nodes. The splitting process is repeated until stopping rules are met i.e. when the class value in the partition is same or there is only one object in the partition. The output of CHAID prediction model is displayed in hierarchical tree-structured form, in which the root is the population, and the branches are the connecting segments such that the variation of the response variable is minimized within all the segments, and maximized among all the segments.

An essential step in CHAID prediction model construction is selecting the relevant features for classification [13]. The purpose of feature selection techniques helps in reduction of computation time and enhances the predictive accuracy of the model. Chi-square [14] is the common statistical test that measures divergence from the distribution expected if one assumes the feature occurrence is actually independent of the class value. Feature Selection via Pearson chi-square ($\chi^2$) test is a very commonly used method [14] and it evaluates features individually by measuring their chi-squared statistic with respect to the classes.

The present investigation used data mining as a tool with CHAID classification tree as a technique to design the student performance prediction model. Filtered feature selection technique [14] was used to select the best subset of variables on the basis of the values of chi-square measure.

## 4. Results

In the present study, those features whose chi-square values were greater than 100 were given due considerations and the highly influencing variables with high chi-square values have been shown in Table 2. These features were used for the CHAID prediction model construction. For both variable selection and CHAID prediction model construction, we have used data mining component of STATISTICA 7.

Table 2: High Potential Variables

| Name of the Variable | Chi-Square Values |
| --- | --- |
| XMARK-Grade | 480.2589 |
| LOC-SCH | 408.4542 |
| MEDU | 345.4228 |
| FEDU | 344.5857 |
| MED | 281.3894 |
| StMe | 276.2699 |
| FSAL | 250.0365 |
| PTuition | 235.2862 |
| FOCC | 182.1662 |
| MSAL | 173.1565 |
| MOCC | 158.7115 |
| LArea | 132.8951 |
| Veh-Home | 123.8983 |
| FAM-Size | 112.4325 |

A tree-based CHAID prediction model for the student performance was constructed using CHIAD algorithm that is shown in Fig. 1. Each node in Fig. 1 contains the details of *node-id* (ID), *number of data objects* (N), and *the possible outcome of the class variable* (HScGrade) with high probability. There are 11 terminal nodes, resulting from 5 *if-then* conditions to predict the performance of the students. The tree starts with the top decision node (ID=1) with 772 instances of the data set and the whole data set is divided into two partitions based on the values of splitting predictive variable MED ("*Tamil*", "*English*"). The left most node (ID=2) with 433 *Tamil* medium students shows a majority of cases associated with *C*-grade students, whereas the right most node (ID=3) contains 339 students of *English* medium shows a majority of *B*-grade students. The leftmost node (ID=2) containing 433 instances is further split on the basis of the value of predictor variable-LArea (*corporation*, *municipal*), resulting in two more nodes (node ID=4 and node ID=5) and so on. Similarly the rightmost node (ID=3) containing 433 instances is further split based on the predictor variable- XMARK-Grade values (*B*, *A*) that results in two more nodes (node ID=6 and node ID=7) each containing 85, 254 instances respectively. The splitting process is explored on both sides further, until either split does not help to improve the predictive accuracy or a node contains instances which are less than the pre-defined size.

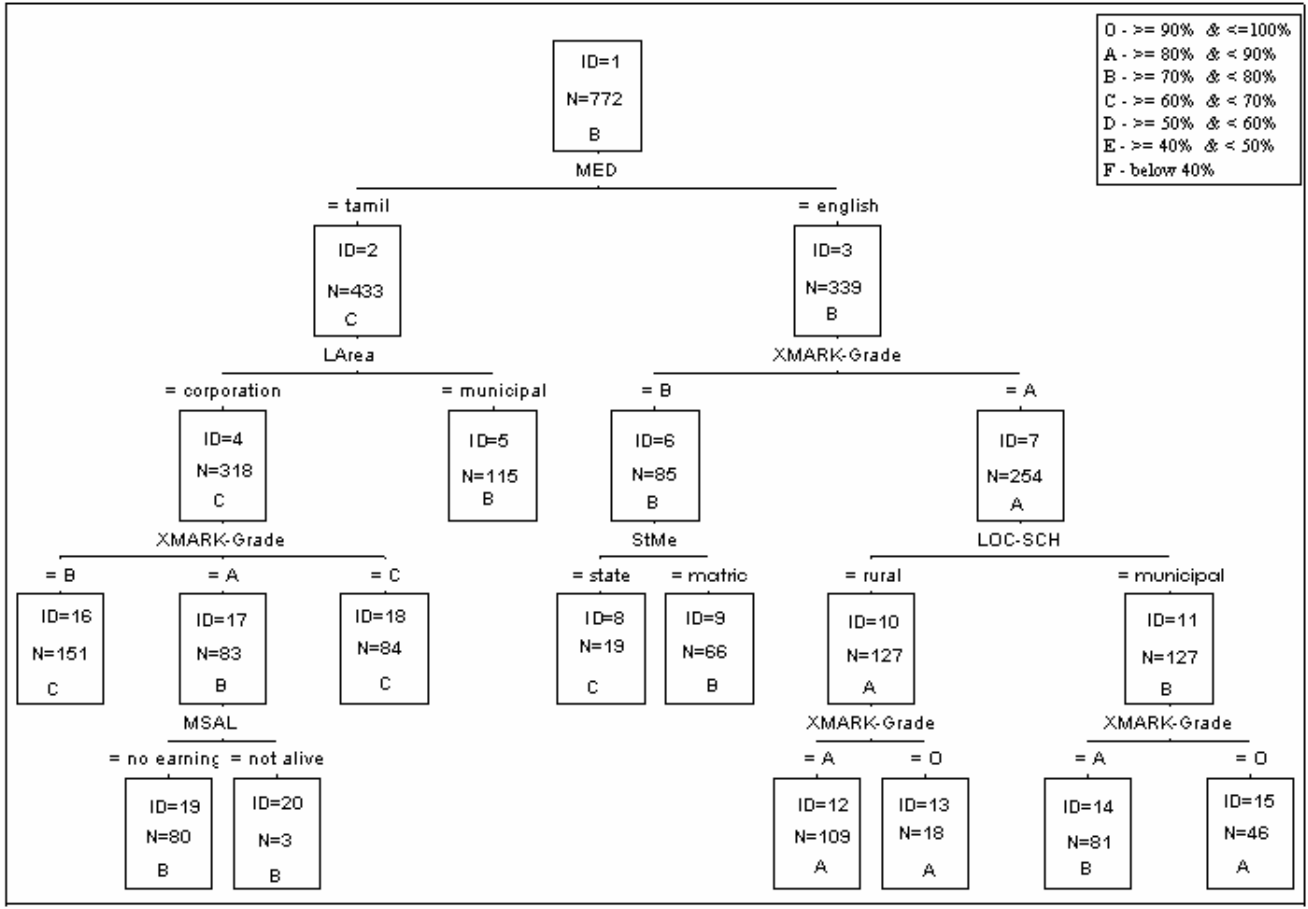

Fig. 1 CHAID Tree classification model.

## 4.1. Rules Extracted from CHAID Tree

The classification rules can be generated by following the path from each terminal node to root node. Pruning technique was executed by removing nodes with less than desired number of objects.

Table 3: Rule Set generated by CHAID Tree

| Rules for HScGrade='A' |
|---|
| IF MED = 'English' and XMARK-Grade = 'A' or 'O' and LOC-SCH = 'rural' and XMARK-Grade = 'A' THEN HScGrade ='A' |
| IF MED = 'English' and XMARK-Grade = 'A' or 'O' and LOC-SCH = 'rural' and XMARK-Grade = 'O' THEN HScGrade ='A' |
| IF MED ='English' and XMARK-Grade='A' or 'O' and LOC-SCH = 'municipal' and XMARK-Grade = 'O' then HScGrade ='A' |

| Rules for HScGrade='B' |
|---|
| IF MED = 'Tamil' and LArea = 'municipal' or LArea = 'rural' then HScGrade='B' |
| IF MED = 'Tamil' and LArea = 'corporation' and XMARK-Grade = 'A' or 'O' and MSAL = 'no earnings' or 'less than 3000' or 'above 15000 and less than 20000' THEN HScGrade='B' |
| IF MED = 'Tamil' and LArea = 'corporation' and XMARK-Grade = 'A' or 'O' and MSAL = 'not alive' or MSAL ='above 5000 and less than 10000' THEN HScGrade='B' |
| IF MED = 'English' and XMARK-Grade = 'B' or 'C' and StMe = 'matric' THEN HScGrade ='B' |
| IF MED = 'English' and XMARK-Grade = 'A' or 'O' and LOC-SCH = 'municipal' and XMARK-Grade = 'A' THEN HScGrade='B' |

| Rules for HSCGrade='C' |
|---|
| IF MED = 'Tamil' and LArea = 'corporation' and XMARK-Grade = 'B' or 'D' THEN HScGrade='C' |
| IF MED = 'Tamil' and LArea = 'corporation' and XMARK-Grade = 'C' or 'E' THEN HScGrade='C' |
| IF MED = 'English' and XMARK-Grade = 'B' or 'C' and StMe = 'state' THEN HScGrade='C' |

Subsequently, the 10-fold cross method for the validation of the model was applied during CHAID prediction model construction process. It was noticed that the decision tree possessed 11 terminal nodes, which resulted in 11 classification rules. These rules have been presented in Table 3.

From the rule set for HScGrade = 'A', it was found that medium of instruction and previous academic achievement at secondary level had influence on academic achievement in higher secondary level. The *English* medium students maintained their academic performance both at secondary and higher secondary level despite of location of school (LOC-SCH), which had good correlation with the previous findings of Ramasamy [16]. The inference from the rule set for HScGrade ='C' showed that the performance of the students with C grade in secondary level did not show any improvement in the higher secondary level. The rule set given in Table 3 focuses only on the three classes namely *A*, *B* and *C*, since the number of objects in these three classes was found to be more. As the number of objects in the rest of the classes was less, they were not generated in the rule set. The classification matrix has been presented in Table 4, which compared the actual and predicted classifications. In addition, the classification accuracy for the seven-class outcome categories was presented.

Table 4: Classification Matrix- CHAID Prediction Model

| HScGrade class values | | predicted | | | | | | % of correct prediction |
|---|---|---|---|---|---|---|---|---|
| | | O | A | B | C | D | E | F | |
| observed | O | | 33 | 4 | | | | | 0.00 |
| | A | | 90 | 61 | 8 | | | | 56.60 |
| | B | | 46 | 149 | 62 | | | | 58.00 |
| | C | | 4 | 87 | 105 | | | | 53.60 |
| | D | | | 21 | 46 | 1 | | | 0.00 |
| | E | | | 1 | 2 | | | | 0.00 |
| | F | | | 21 | 31 | | | | 0.00 |

It was found that the overall model prediction accuracy of CHIAD prediction model was 44.69% and it indicated that the CHAID model could correctly classify 345 students among 772 students. The accuracy of the present model was compared with other models and it was found to be higher than the accuracy of earlier model that was generated by Al-Radaideh et al. [10].

## 5. Conclusions

The CHAID prediction model was useful to analyze the interrelation between variables that are used to predict the outcome on the performance at higher secondary school education. The features like medium of instruction, marks obtained in secondary education, location of school, living area and type of secondary education were the strongest indicators for the student performance in higher secondary education. This CHAID prediction model of student performance was constructed with seven class predictor variable, whereas the earlier models in reviews were constructed with limited class predictor variables.

Even though CHAID model handled small and unbalanced data set, it could be worked out effectively with better predictive accuracy. By applying Boosting and Bagging, which are two predominant techniques, the predictive accuracy would be further improved. Generalization of the outcomes could not be made due to the limited samples of students in limited geographical coverage of schools in different districts of the state of Tamilnadu. A hybrid model, which utilizes the full impact of variable selection and its consequences, is being worked out and it is expected to present a better picture of the real scenario.

**M. Ramaswami** is Associate Professor in the Department of Computer Science, Madurai Kamaraj University, TamilNadu, India. He obtained his M.Sc. (Maths) from Bharathiyar University, M.C.A from Madurai Kamaraj University and M.Phil. in Computer Applications from Manonmaniam Sundaranar University. He is currently doing research in Data Mining and Knowledge Discovery.
**R. Bhaskaran** received his M.Sc. (Mathematics) from IIT, Chennai, India (1973) and obtained his Ph.D. in Mathematics from the University of Madras (1980). He then joined the School of Mathematics, Madurai Kamaraj University as Lecturer. At present, he is working as Professor of Mathematics. His research interests include Non-Archimedean analysis, Image Processing, Data Mining and Software development for learning Mathematics.